\documentclass{article}
\usepackage{graphicx}
\usepackage{latexsym}
\usepackage{CJKutf8}
\usepackage{amsmath}
\usepackage{url}

\begin{document}

\title{Towards Bridging the Digital Language Divide}
\author{Gábor Bella\thanks{University of Trento, Italy.}, Paula Helm\thanks{University of Amsterdam, The Netherlands.}, Gertraud Koch\thanks{University of Hamburg, Germany.}, and Fausto Giunchiglia\thanks{University of Trento, Italy.}}
\date{}

\maketitle

\begin{abstract}
It is a well-known fact that current AI-based language technology---language models, machine translation systems, multilingual dictionaries and corpora---focuses on the world's 2--3\% most widely spoken languages. Recent research efforts have attempted to expand the coverage of AI technology to `under-resourced languages.' The goal of our paper is to bring attention to a phenomenon that we call \emph{linguistic bias}: multilingual language processing systems often exhibit a hardwired, yet usually involuntary and hidden representational preference towards certain languages. Linguistic bias is manifested in uneven per-language performance even in the case of similar test conditions. We show that biased technology is often the result of research and development methodologies that do not do justice to the complexity of the languages being represented, and that can even become ethically problematic as they disregard valuable aspects of diversity as well as the needs of the language communities themselves. As our attempt at building diversity-aware language resources, we present a new initiative that aims at reducing linguistic bias through both technological design and methodology, based on an eye-level collaboration with local communities.
\end{abstract}

\section{Introduction}

The notion of \emph{digital language divide} refers to the gap between languages with and without a considerable representation on the Web and within the worldwide digital infrastructure \cite{divide}. As shown by \cite{kornai2013digital} about 10~years ago, less than~5\% of the world's 7--8,000 languages have an even remotely significant representation on the Internet. The same orders of magnitude remain valid today, despite the progresses of a decade. Due to the inextricable link between language, culture, and society (as we show through many examples in this paper), the ability of persons and peoples to express themselves in their own language is determinant in maintaining their identity and their unique perspective in which ideas and worldviews are anchored, and which are thus crucial for the dignity of human beings \cite{Nyabola_2018, UNESCO.2005}.

In the field of language technology and research, riding the wave of the recent breakthrough of neural AI, the last decade saw a surge in multilingual language tools and resources for `under-resourced languages.' Researchers have tried to enable technologies such as machine translation, natural language processing, or speech recognition, to an ever larger scope of languages. However, it has been pointed out that many such efforts are based on a simplistic vision according to which, with the help of AI already successfully applied to well-resourced languages, first-world experts will bring forth solutions that `save the day' in the Global South \cite{bird-2020-decolonising,schwartz-2022-primum}. This attitude leads to a misalignment between the interests and solutions of the former and the real-world needs of the latter. Worse, due to a general ignorance of the languages and, ultimately, the cultures being worked upon---a modus operandi that has become the norm due to the data-driven nature of statistical and neural AI---major quality problems in the results go unnoticed \cite{lignos-etal-2022-toward}.

The first of the three contributions of our paper is to draw attention to the deep, meaning-level, and so far under-investigated phenomenon of \emph{linguistic bias} in technology. A language resource or tool manifests linguistic bias if, due to its design, it is unable to represent or process adequately the language to which it is applied. Linguistic bias is tightly related to a second key notion of our paper, \emph{linguistic diversity}, that refers to linguistic features and ideas that are `hard to translate' across languages. 
%
Our second contribution is to show that the technological reasons of linguistic bias are, in part, a consequence of flawed methodologies in the way technology is currently built. 
We argue that one of the main requisites for a diversity-aware, low-bias technology design is an engaged participation of local communities, via a local institutional framework, whenever possible.
As our third contribution, we present the case study of the \emph{LiveLanguage} initiative, our attempt at building large-scale multilingual lexical resources that are diversity-aware by design and in which we try to reduce linguistic bias systematically. We present a diversity-aware lexical data model and a collaborative methodology through which the contents of LiveLanguage have been extended during the last years. Yet, we are aware of the fact that the uneven digital representation of languages is a complex problem set, of which linguistic diversity and bias are but a puzzle piece. Therefore, we consider the solutions provided here as one necessary, but by no means a sufficient condition for bridging the digital language divide.

The rest of the paper is organised as follows. Section~2 defines and discusses the notions of linguistic diversity and bias. Section~3 provides examples of bias from AI-based language technology. Section~4 provides a critique of methodologies in language technology research and development, and proposes more ethical alternatives. Sections~5 and~6 describe the technical and methodological aspects of our LiveLanguage initiative, aiming to put into practice the principles discussed in Sections~2--4. Finally, Section~7 situates the current state of our work with respect to long-term goals.

\section{Linguistic Diversity and Bias}
\label{sec:bias}

The term \emph{linguistic diversity} has a broad positive connotation: evocative of \emph{biodiversity}, its association to language implies the preservation of the variedness of the world's linguistic landscape. Although our own point of departure is indeed one of 
preserving diversity, we are wary of naïvely celebrating it without a proper conceptualization \cite{Helm_Michael_Schelenz_2022}. Therefore, we differentiate between an understanding of linguistic diversity as a \textit{descriptive} and as a \textit{normative} concept, to better distinguish between (a)~the actual notions of difference that underlie our understanding of linguistic diversity as a design strategy in computational systems, and (b)~the values we associate with diversity as the objective of our work. 

As a descriptive notion,   \cite{greenberg1956measurement} defined linguistic diversity as the probability of two persons speaking the same language in a certain geographic area. \cite{rijkhoff1993method}, instead, applies the term (informally) to sets of languages, and understands the `variedness' of the languages in terms of their genetic relationships, an idea also adopted in \cite{giunchiglia2017understanding} as a measurement of the genetic diversity of a set of languages. 
On the normative side, diversity can be seen as a value that is either intrinsic or instrumental \cite{ZimmermanBradley.2019}. In the former case, diversity is good by and for itself, and evokes romantic associations with pluralism, tradition, and authenticity \cite{Vertovec.2012,Chi.2021}. 
It is, however, also important to acknowledge the limits of the value of diversity.  Trade languages such as English, for example, spoken by various peoples across the globe, enable mutual understanding and exchange of ideas.  

Given these conceptual clarifications, we embrace linguistic diversity as an objective, together with the idea that computational efforts can be instrumental to achieve it. 
In this perspective, we first provide a general and descriptive definition for linguistic diversity, followed by an operational and normative interpretation of what dealing with linguistic diversity implies in practice.

\begin{quotation}\textit{ \textbf{Linguistic diversity} is observed across two (or more) languages if one language possesses a particular linguistic feature through which it can express an idea concisely, while the same feature is absent from the other language that, in consequence, needs to express it through different features, if at all.}
\end{quotation}
This definition is very general, with the term \emph{linguistic feature} possibly referring to any lexical, grammatical, etc.~phenomenon; yet, the reference to \emph{expressing an idea} implies that the feature in question contributes to the meaning of the utterance in which it is used. We provide here two examples from the field of lexical semantics. The first one is lexical untranslatability: for instance, the Maori word \emph{teina} means \emph{elder brother} if it is pronounced by a male speaker, and \emph{elder sister} if it is pronounced by a female. When translating to English, the gender of the speaker cannot be expressed in this succinct way: it may be communicated through the context, or omitted. Another example is the linguistic feature of \emph{inalienable possession}, widely present in Native American and Australasian languages, where abstract---yet for us natural---concepts such as \emph{mother} or \emph{head} (as a body part) cannot be expressed as single words (free morphemes), but only together with their possessor (i.e.~as the combination of two bound morphemes): \emph{my mother}, \emph{your head}.

For native speakers, the ideas expressed by such language-specific features are often inextricably embedded in the local natural or cultural context. For a speaker in South India, choosing the correct term out of 16~possible terms to designate one's cousin---depending on gender, age, the mother's or father's side, etc.---is a basic requirement of politeness and culture. 
In the Italian Alps, the word \emph{malga}, designating a typical mountain restaurant with no equivalent outside the Alpine region, is an important everyday term with a strong local connotation.

In the context of AI language technology, the notion of \textit{bias} has so far been used to refer to patterns of stereotypes and preferences towards social groups, most often concerning learning-based language processing systems \cite{blodgett2020language}. In terms of social groups, studies have mostly focused on gender, ethnicity, and race, but also other forms of bias (religion-related, age-related, political, etc.)  \cite{FriedmanNissenbaum1996}. To our knowledge, the term \emph{linguistic bias} has not been used so far in any way similar to ours. Many of the underlying exploitative mechanisms have, however, been pointed out, in particular in relation to the most disempowered social groups, namely small indigenous speaker communities \cite{bird-2022-local, schwartz-2022-primum}. In terms of actual bias in AI systems and data, the research closest to ours concerns \emph{inductive bias} in language models towards certain morphological and syntactic structures \cite{ravfogel-etal-2019-studying,white-cotterell-2021-examining}. We present these works more in detail in Section~\ref{sec:sources_of_bias}. In \cite{blodgett2016demographic}, Blodgett et al.~study the (non-)representation of the vernaculars of social groups within language resources. They point out that English linguistic corpora tend to exclude the register of speech used by African-Americans, the non-representation of which causes a bias in the abilities of the AI systems trained on top of them. We identify this as a particular case of linguistic bias, even if in the paper cited, it is (also correctly) framed as a form of racial bias within the context of a single language. While our definition of linguistic bias below aims to encompass such cases, the focus of our current research is bias across languages rather than across registers, styles, or vernaculars within the same language.

Intuitively, linguistic bias is observed in language technology when, \emph{due to its design}, a system represents, interprets, or processes utterances in certain languages less precisely or less efficiently than in others, thereby negatively affecting the communication ability of speakers of that language. More formally:
\begin{quotation}
A technology~$t$ that supports the languages $L=\{l_1,...,l_N\}$ is \textbf{linguistically biased} if there exist a pair of languages $l_A\in L,l_B\in L$, an operation~$o_t$ performed by $t$, a set of utterances~$U_A$ in language~$l_A$ given as input to~$o_t$, and a set of input utterances~$U_B$ in language~$l_B$ equivalent in meaning to $U_A$, such that the performance of $o_t$ over~$U_A$ is distinctly better than its performance over~$U_B$: $\textrm{Perf}(o_t(U_A))>>\textrm{Perf}(o_t(U_B))$.
\end{quotation}
In order to obtain a quantitative measure of linguistic bias~$b$, we use \emph{sample standard deviation} over the performance values measured across $N$ languages:
$$ b_t=\sqrt{\frac{1}{N-1}\sum_{i=1}^{N}{\textrm{Perf}(o_t(U_i))^2-\overline{\textrm{Perf}}^2}} $$
where $\overline{\textrm{Perf}}$ is the mean performance over the languages. 
%
%
%
%
We use the term \emph{performance} in a broad sense, as it can be measured in many ways depending on the task being evaluated: for accuracy, precision, or recall, $0\le\textrm{Perf}\le 1$, the higher the better, while for perplexity or semantic distance there is no fixed upper bound, and the lower the better. 

If applied unconditionally, the definition above will find bias everywhere, as in practice no technology performs equally well on any two language. As most language processing systems today are data-driven, any difference in training data size or quality will inevitably lead to uneven performance. Rather than to reveal such trivial truths, we aim to use linguistic bias to examine deeper, structural limitations built into language processing algorithms, representational models, resources, or methodologies. 
In practice, however, from merely observing the output of a system, it may be difficult to understand whether lower performance is caused by its structural properties or by contingent factors such as resource completeness or training data size. The distinction is important as the latter kinds of issues can be mitigated simply by adding more data to the system, while structural bias can only be addressed by redesigning it. In order to focus on the structural sources of bias, a careful selection of the set of input utterances~$U$ may be necessary, and the systems may even need to be retrained or repopulated with corpora balanced across languages. In the next section, we illustrate through examples how such a balanced measurement of linguistic bias can be carried out in practice.

We recognise that bias is generally unavoidable. 
It is now accepted in humanities and social science research that all knowledge, all insights, and even all data are situated, i.e.~they always reflect a particular point of view in space and time that is influenced by culture, history, politics, economics, epistemology, and so on \cite{Haraway.1988, Gitelman_2013}. 
It is therefore important to be upfront about when and for what reasons a certain bias is problematic and needs to be combated, and that this combating does usually not lead to no bias, but to a different, more transparent and just bias \cite{Harding_1995}. This is the case, for example, when bias targets already vulnerable, underrepresented, and marginalized groups. 

In our case, the social group in question is clearly the community of speakers of a given language, however heterogeneous it may be otherwise (according to social status, culture, gender, race, ethnicity, religion, etc.). Being the native or second-language speaker of a language in itself determines one's access to information, and the language technology that enables this access affects one's ability to communicate, on the Web or elsewhere.


\section{Bias in Language Technology}
\label{sec:sources_of_bias}

This section presents examples of linguistic bias in mainstream AI language technology: within the structure of multilingual lexical databases, within neural language models, and finally various manifestations of linguistic bias in machine translation systems.


\begin{figure}
    \centering
    \includegraphics[width=.7\textwidth]{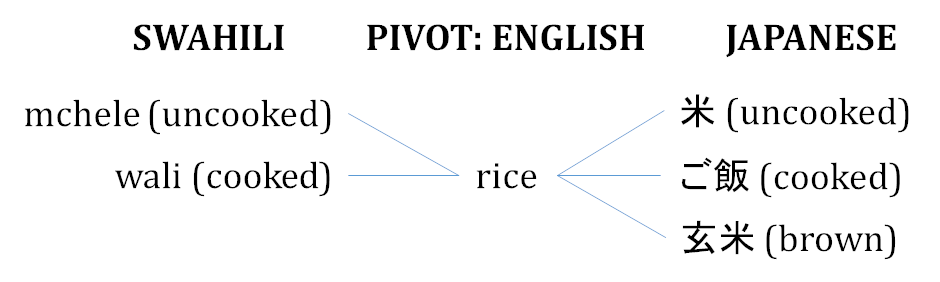}
    \caption{Biased cross-lingual mapping of words about various forms of `rice' from a popular multilingual lexical database.}
    \label{fig:mldb_mapping}
\end{figure}

\begin{figure}
    \centering
    \includegraphics[width=\textwidth]{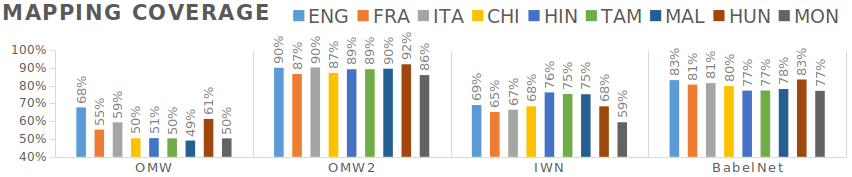}
    \caption{Bias in the expressivity of cross-lingual mappings of multilingual lexical databases. A more uniform per-language coverage means a lower bias.}
    \label{fig:bias_mldb}
\end{figure}

\begin{figure}[t]
    \centering
    \includegraphics[width=.9\textwidth]{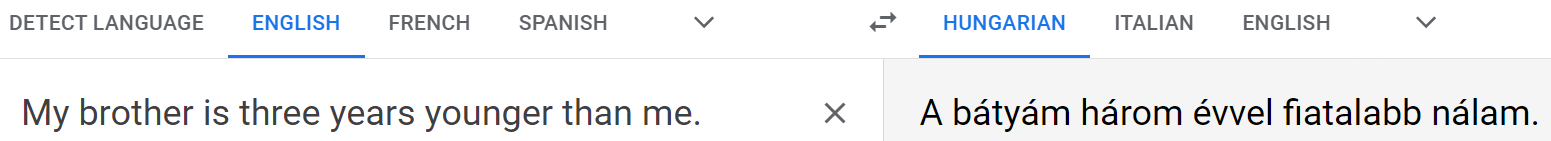}\\
    (a)\\
    \includegraphics[width=.9\textwidth]{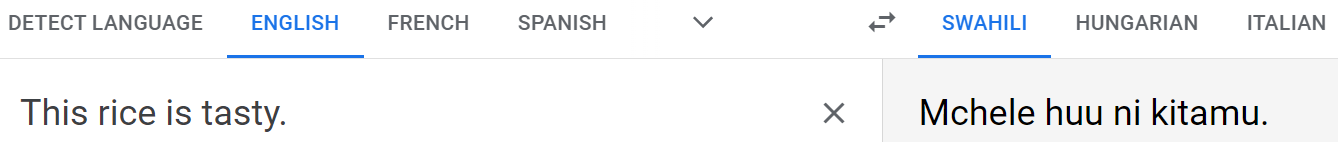}\\
    (b)\\
    \includegraphics[width=.9\textwidth]{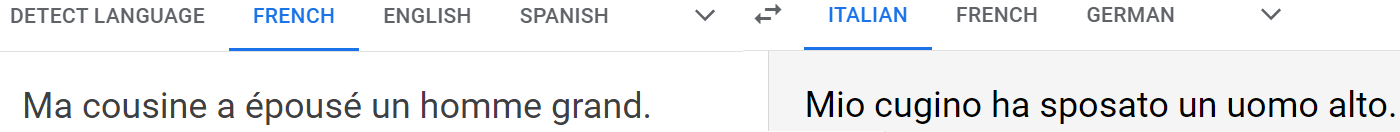}\\
    (c)\\
    \caption{Examples of linguistic bias in machine translation.} 
    \label{fig:mt}
\end{figure}


\paragraph{Bias in Lexical Databases.} As a generalisation of bilingual dictionaries, the 2000s saw the appearance of \emph{multilingual lexical databases} that map words, based on their meanings, across large numbers of languages. While these resources proved to be extremely useful in cross-lingual applications, looking under the hood---into their underlying models of lexical meaning---reveals varying levels of limited expressivity and bias.

As shown in the survey \cite{giunchiglia2023representing}, several of these multilingual databases interconnect words from hundreds of languages, mapping the words of each language to the 100~thousand English word meanings (so-called \emph{synsets}) of Princeton WordNet \cite{miller1998wordnet}. On the one hand, this choice makes practical sense, as among all similar resources, WordNet offers by far the widest coverage of word meanings. On the other hand, it results in a strong bias towards the English language and Anglo-Saxon culture in general, as the expressivity of the database is limited to notions for which a word exists in English \cite{giunchiglia2023representing,bella2022linguistic}. Figure~\ref{fig:mldb_mapping} provides a simple example from the food domain, known to be culturally, and thus also linguistically, diverse. It shows how a biased lexical database maps together words in Swahili and Japanese meaning \emph{uncooked rice}, \emph{cooked rice}, and \emph{brown rice}. The degree of information loss is flagrant: while both Swahili and Japanese provide fine-grained lexicalisations about the various forms of rice, the many-to-many mapping that results from passing through English masks all fine-grained differences, resulting in both a loss of detail and incorrect translations when one moves from Swahili to Japanese or vice versa. The diversity-diminishing bias towards the English language and Anglo-Saxon cultures is also found in other domains that are well-known to be diverse across languages: family relationships, school systems, etc.

Applying the definition from Section~\ref{sec:bias}, we compute the linguistic bias of the lexical models of four multilingual lexical databases: the first and second versions of the Open Multilingual Wordnet (OMW, OMW2) \cite{bond2013linking,bond-etal-2020-issues}, IndoWordNet (IWN) \cite{bhattacharyya2010indowordnet}, and BabelNet (BN) \cite{navigli2010babelnet}. The figures were not computed from the actual contents of the databases---that are necessarily incomplete and thus are not representative of their structural properties---but rather from synthetic data and a gold-standard cross-lingual mapping dataset, both obtained from  \cite{giunchiglia2023representing}. Coverages are theoretical in the sense that they represent the percentage of cross-lingual mapping relationships that each model is able to represent with respect to the gold standard mappings. The latter consist of three mapping relation types: \emph{equivalent meaning}, \emph{broader/narrower meaning}, and \emph{lexical untranslatability}.
\begin{itemize}
    \item technology $t$: multilingual lexical databases \{OMW, OMW2, IWN, BN\};
    \item operation $o_t$: cross-lingual mapping of word meanings;
    \item languages $L$: \{English, French, Italian, Chinese, Hindi, Tamil, Malayalam, Hungarian, Mongolian\};
    \item utterances $U$: 32~concepts from six linguistically diverse domains, lexicalised by the languages above as 160~word senses and 128~lexical gaps;
    \item performance Perf: defined as the coverage (recall) of gold-standard cross-lingual mappings that the lexical database is able to express.  
\end{itemize}
Figure~\ref{fig:bias_mldb} shows the results. We are not concerned with absolute coverage values but rather with how coverage varies across languages.
OMW shows a marked bias towards European languages and English in particular (68\% coverage) while Asian languages are mapped suboptimally (49--51\%), with a bias of $b_\textrm{OMW}=6.53\%$. This is explained by the fact that pivot concepts in OMW are limited to the meanings of English words. IWN, where the pivot is Hindi, unsurprisingly displays a bias towards Indian languages (75--76\%) with other languages falling into the 59--68\% range, with $b_\textrm{IWN}=5.51\%$. BN and OMW2, on the other hand, are less biased due to the fact that their pivot concepts are not tide to any particular language. The bias of BN, $b_\textrm{BN}=2.52\%$, the mapping coverage of which varies between 77\% and 83\%, is due to its lack of support for untranslatability relations. In the case of OMW2 (coverage 86--92\%), the so far smallest bias $b_\textrm{OMW2}=1.96\%$ is caused by limited expressivity in cross-lingual broader/narrower mappings.


\paragraph{Bias in Neural Language Models.}
Do neural language models favour, in terms of better performance, certain types of languages based on their grammatical features (e.g.~word order, morphology)? This topic has not been widely researched, perhaps due to the variety of typological differences across languages, which makes it hard to examine the impact of a single linguistic feature in isolation. In order to circumvent this problem, and in line with our intuition in Section~\ref{sec:bias} that the evaluation of structural bias requires abstraction from contingent factors, \cite{ravfogel-etal-2019-studying} and \cite{white-cotterell-2021-examining} both construct artificial languages that differ by single typological features. Apart from this similarity, the two studies are different: the former examines the behaviour of a single LSTM-based architecture over three main types of grammatical features (agreement, word order, case) over a modified version of English. The latter compares LSTM and Transformer architectures trained over entirely artificial languages generated by systematic variations of a relatively small context-free grammar, in terms of word order only. Using our formalism, the setup of the latter can be characterised as follows:
\begin{itemize}
    \item technology~$t$: language models \{LSTM, Transformer\};
    \item operation~$o_t$: word prediction;
    \item languages $L$: 64~artificial languages reproducing word order features that appear in natural languages;
    \item utterances~$U$: 100k sentences constructed through artifical CFGs;
    \item performance Perf: perplexity (PP, the higher the less accurate).
\end{itemize}
\cite{white-cotterell-2021-examining} found LSTM to be insensitive to change in word order ($22<\textrm{PP}_\textrm{LSTM}<24$) and Transformer, on the contrary, to be sensitive to it ($25<\textrm{PP}_\textrm{Tr}<37$). In particular, the Transformer was found to favour the rarely-occurring VOS (verb--object--subject) order (3\% of all languages), but no major difference was found between the most frequent SOV and SVO orders.  

\paragraph{Bias in Machine Translation.} Machine translation (MT) has been flagship task of AI-based language technology. 
Without claiming to be exhaustive, we point out three aspects of current MT technologies where lingusitic bias can be observed: the non-handling of untranslatability, the variedness of vocabulary and grammar, and the use of a pivot language.

Today's top MT systems, such as DeepL and Google Translate, make systematic mistakes over untranslatable terms, betraying the fact that this phenomenon is not specifically addressed by these tools \cite{khishigsuren2022using}. The screenshots~(a) and~(b) in Figure~\ref{fig:mt}, taken from a mainstream machine translator, show examples of erroneous translations due to untranslatability. For instance, when translating the English sentence \textit{My brother is three years younger than me} to Hungarian, Korean, Japanese, or Mongolian syntactically correct yet semantically absurd results are obtained.
These languages either have no equivalent word for \emph{brother} or, when they do, the equivalent word is relatively rare (as \emph{fiútestvér} in Hungarian). Based on training corpus frequencies, the MT system ends up choosing a semantically unsuitable word, such as \emph{bátyám} meaning \emph{my elder brother}, resulting in \emph{My elder brother is three years younger than me.} A similar example, based on the example of \emph{rice} from earlier in this section, is the English sentence \emph{`This rice is tasty,'} machine-translated into Swahili as \emph{`Mchele huu ni kitamu,'} meaning \emph{`This raw rice is tasty.'} These are not cherry-picked exceptions but rather are examples of systematic mistakes from domains of high linguistic diversity.

Given the nature of the errors above---the use of words with incorrect meanings within otherwise syntactically correct sentences---and taking inspiration from \cite{khishigsuren2022using}, we propose a measurement of bias based on lexical semantics. We use as a measurement of performance the average semantic distance, more precisely the \emph{least common subsumer distance}, between the meaning of each translated word and the expected gold-standard meaning, measured over the interlingual concept hierarchy published in \cite{khishigsuren2022using}.
\begin{itemize}
    \item technology~$t$: Google Translate \{GT\};
    \item operation~$o_t$: translation from English;
    \item languages $L$: Russian, Japanese, Korean, Hungarian, Mongolian;
    \item utterances~$U$: 50~English sentences from the British National Corpus containing kinship terms, from \cite{khishigsuren2022using};
    \item performance Perf: average semantic distance (the lower the better) between the meanings of translated words and correct gold-standard concepts.
\end{itemize}
The average semantic distances obtained, as reported in \cite{khishigsuren2022using}, are $\textrm{Perf}(\textrm{GT}_\textrm{ENG--RUS})=0.34$, $\textrm{Perf}(\textrm{GT}_\textrm{ENG--JAP})=0.38$, $\textrm{Perf}(\textrm{GT}_\textrm{ENG--KOR})=0.90$, $\textrm{Perf}(\textrm{GT}_\textrm{ENG--HUN})=1.06$, $\textrm{Perf}(\textrm{GT}_\textrm{ENG--MON})=1.12$, which provides an overall bias of $b_\textrm{GT}=0.374$ (to be compared with the biases of other MT systems). 

A second form of bias concerns the variedness of vocabulary and grammar in MT output. In \cite{vanmassenhove-etal-2021-machine}, Vanmassenhove et~al.\ quantitatively compare the lexical and grammatical diversity between original and machine-translated text. Their definitions of diversity and bias are different from ours: by diversity they refer to the richness of the vocabulary and the complexity of the grammar of a document (normalised by document size and computed according to various metrics), while by bias they understand an uncontrolled loss of diversity due to MT. Still, their results are relevant under our interpretation: \cite{vanmassenhove-etal-2021-machine} reports that, for the same language, morphology in translated text becomes poorer with respect to original (untranslated) corpora, i.e.~features of number or gender for nouns tend to decrease. This phenomenon affects morphologically rich languages in particular. 

A third form of bias in MT systems is their use of English as a pivot language when translating between non-English language pairs. This practice is explained by the relative scarcity of bilingual training corpora for such language pairs, as well as scalability: the use of a pivot language reduces the need for trained models from $\binom{N}{2}$ to $N-1$, where $N$ is the number of languages. Example~(c) in Figure~\ref{fig:mt} shows the case of French-to-Italian translation of a sentence meaning \emph{my (female) cousin married a tall man.} While French and Italian use different words for male and female cousins (\emph{cousin/cousine}, \emph{cugino/cugina}), English does not. The result is that the gender of the cousin is `lost in translation' and, as a form of combined linguistic and gender bias, it appears as a male in the translated text.

\section{Methodology as a Source of Bias}

In our view, bias in language technology exists in a large part due to methodological flaws in computational linguistics research and development practices. 
In Computational Linguistics, English has not only been the lingua franca of scientific communication, but also the de facto standard subject matter of research. 
\cite{schwartz-2022-primum} reports that between 2013 and 2021, 83\% of papers accepted at ACL were explicitly or implicitly about English and 97\% were about Indo-European languages. 
The 2010s saw an emerging interest in multilingual language technology, and of a new research sub-field targeting `low-resource' (or `under-resourced') languages, previously neglected by mainstream research. 
%
Nevertheless, in line with the `zero-shot' data-driven ethos \cite{bird-2020-decolonising} of recent deep AI research that shuns any use of prior results from linguists and field workers, low-resource language research is only worthy of a top publication as long as 
(1)~it provides a solution for multiple, preferably tens or hundreds of languages at the same time; 
(2)~it involves mainstream AI technology, i.e.~neural networks; and 
(3)~it requires very little to no knowledge from authors and readers of the languages targeted. 
%

The methodology and the goals of such research raise a multitude of ethical concerns. 
Linguists such as Bird \cite{bird-2020-decolonising} 
claim mainstream Western `low-resource language' research to be postcolonial, with Western researchers unilaterally setting developmental goals and providing technological solutions to reach them. Bird points out how typical research agendas such as language documentation and `technology-based revitalisation' do not respond to the needs of local communities. Most often, native speakers are not involved in the process, or when they are, they play subordinate roles such as annotator or validator. In many cases, the languages being addressed are not even remotely understood by the people working on them, who are therefore not able to judge the quality of the data they are relying on, and sometimes not even whether they are using the right language, as in the case of automated Wikipedia scraping \cite{lignos-etal-2022-toward}.
These criticisms are in line with the challenges identified in relation to what Irani et al.\ describe as `postcolonial computing,' and with the 4-dimensional ways forward that the respective authors propose \cite{Irani_Vertesi_Dourish_Philip_Grinter_2010}. 


\begin{figure*}[t]
    \centering
    \includegraphics[width=\textwidth]{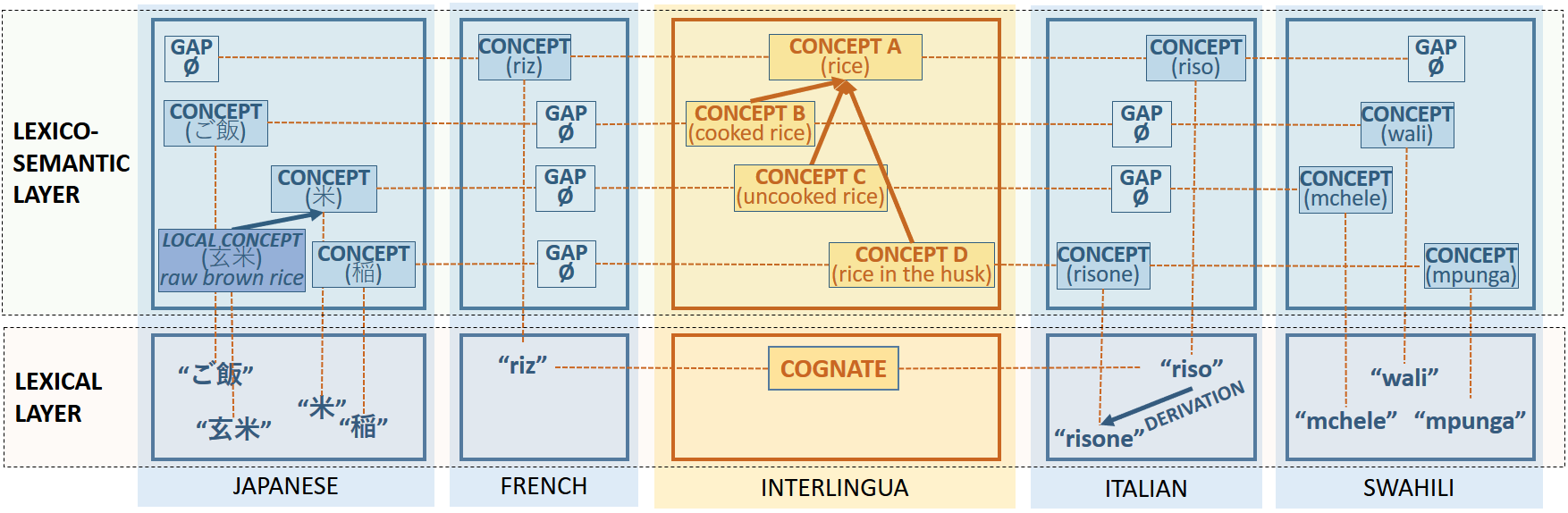}
    \caption{The cross-lingual mapping model of the UKC lexical database}
    \label{fig:ukc_model}
\end{figure*}



In deliberate opposition to the shortcomings described in Section~\ref{sec:sources_of_bias}, a few researchers such as Bird \cite{bird-2020-decolonising} and Schwartz \cite{schwartz-2022-primum} have been advocating alternative, `decolonising' approaches to multilingual research in AI and to working specifically with indigenous linguistic communities, based on an understanding of power imbalance and the difference in epistemologies between the researcher and the local community, and overcoming them through deliberate effort. 
Bird emphasises \emph{co-design} of technology with communities, based on their perceived goals and needs. 
He observes the importance of \emph{vehicular} or \emph{trade languages} in addressing local vernaculars. While English is indeed used as lingua franca in many parts of the world, Arabic and Persian in the Middle East, Hindi/Urdu in Northern India and Pakistan, Hausa and Swahili in Africa, etc., are also widely used trade languages. In \cite{bird-2022-local}, a \emph{multipolar model} is proposed for working with language communities. As trade languages function as bridges across local languages and are widely spoken by local communities, according to the multipolar model, they are better adapted as pivots for describing and interconnecting these vernaculars. 

\begin{table}
    \setlength{\tabcolsep}{2pt}
    \centering
    \begin{tabular}{lrrr||lrrr}
        Language & L1+2 spk & Rank & Articles & Language & L1+2 spk & Rank & Articles \\\hline
English & $>$1B & 1 & 6624314 & Swahili & 80M & 83 & 76,417 \\
Indonesian & 300M & 22 & 639717 & Hausa & 77M & 123 & 21,190 \\
Bengali & 300M & 63 & 134,966 & Pashto & 40M & 127 & 17,202 \\
Marathi & 100M & 74 & 90,421 & Sc.~Gaelic & 50k & 133 & 15,859  \\
Breton & 200k & 82 & 78,361 \\\hline
\end{tabular}
\caption{\label{tab:languages}Contrast of the number of speakers (as first or second language) and the number of Wikipedia articles for a few selected example languages.}
\end{table}

We endorse the idea of a co-design methodology where local communities exercise decisional power and property rights over research outcomes. We embrace Bird's multipolar model, both as a methodological approach and as a high-level system architecture for the development of linguistic knowledge. Our difference with the perspective of Schwartz and Bird lies mainly in the communities and challenges targeted: while these authors adopt the viewpoint of small and disempowered indigenous communities (e.g.~Australian aboriginals, Native Americans), we point out the need for a finer-grained typology of the  linguistic communities targeted, in order to develop an ethical framework that best corresponds to the community at hand, sometimes markedly different from small indigenous groups.
Communities with tens of millions of speakers are rarely disempowered linguistic minorities. As shown in Table~\ref{tab:languages},\footnote{Retrieved in February 2023 from \url{https://meta.wikimedia.org/wiki/List_of_Wikipedias}.} languages such as Bengali, Urdu, or Indonesian are each spoken by 100~million people or more. Swahili, Hausa, or Pashto are each spoken by at least 50~million. Yet, the online presence of these languages is nowhere representative of such numbers.\footnote{Kornai \cite{kornai2013digital} has quantitatively proven a very strong correlation between simple measures such as Wikipedia presence and the general digital vitality of a language. For this reason, we consider the number of Wikipedia pages as a decent estimate for the overall digital content available in a language.} While the largest of these communities may actually see themselves `supported' by Western tech multinationals out of sheer economic interest, such support dwindles away below a certain cutoff point of population size and purchasing power. 

Languages such as Breton or Scottish Gaelic fall into yet another category, that of endangered minority languages of the Global North. 
These languages are characterised by a small number of speakers in steep decline, yet with an economic and socio-cultural support much higher than that of indigenous minorities in other parts of the world. This is also reflected in Table~\ref{tab:languages} where, in terms of Wikipedia content, Breton (200~thousand speakers) is on a par with Swahili (80~million) and Scottish Gaelic (50~thousand) with Pashto (at least 40~million).

What distinguishes the communities described above from the small indigenous ones targeted by Schwartz and Bird is that the former enjoy a non-negligible level of institutional backing: official language status and the corresponding administrative and academic support for the major languages of the Global South, and at least financial aid and academic backing for the minority languages of the Global North. Such existing frameworks of support need to be taken into account when setting up collaborative efforts.


\section{A Diversity-Aware Lexical Model}
\label{sec:solution_technology}

As a case study on diversity-aware language technology, we present the \emph{Universal Knowledge Core} (UKC), a large-scale multilingual lexical database  \cite{giunchiglia2018one,bella2022language}. The lexical model of the UKC was built to allow for the representation of the linguistic and cultural plurality of communities, bridging them in a way that does not structurally favour specific languages \cite{giunchiglia2023representing}.

The UKC adopts a language-agnostic lexical concept space, extensible by concepts that are culturally or linguistically specific to languages and communities, avoiding the type of bias illustrated in Figure~\ref{fig:mldb_mapping}. Unlike other databases, however, the UKC also explicitly represents evidence of linguistic diversity. As we show below, language processing systems can reuse such evidence to reduce linguistic bias in their output.


Figure~\ref{fig:ukc_model} shows the high-level lexical model of the UKC, based on our running example around words related to rice. 
Horizontally, the model is divided into two layers: the \emph{lexico-semantic layer} represents lexical meaning through concepts and their relationships  (broader--narrower, part-of, etc.). The \emph{lexical layer} represents the lexicalisations of these concepts---\emph{word senses} in WordNet terminology---as well as associative relations between them. Vertically, the model is divided into an \emph{interlingua} (in yellow) that models unity, i.e.~shared phenomena across languages, as well as one lexicon per language (in blue) that models diversity. This bidimensional structure delineates four types of lexical knowledge: lexico-semantic unity, lexico-semantic diversity, lexical unity, and lexical diversity.

\emph{Lexico-semantic unity.} In the lexico-semantic interlingua, an interlingual concept graph represents lexical meaning shared across the world's languages (the concepts of \emph{rice}, \emph{cooked rice}, \emph{uncooked rice}, and \emph{rice in the husk} in Figure~\ref{fig:ukc_model}). Interlingual concepts are not \emph{a priori} biased towards any particular language, hence the abstract names \textsc{concept~A, B}, etc.\footnote{Even though these concepts are interlingual, we included in the figure their meanings in English for understandability.} Thus, the interlingua asserts that the French \emph{riz} and the Italian \emph{riso} are equivalent, as both are connected to \textsc{concept~a} of \emph{rice}. Likewise, the Swahili \emph{mchele} and the Japanese \begin{CJK}{UTF8}{min}米\end{CJK} are connected to the interlingual \textsc{concept~c} of \emph{uncooked rice}, which is asserted by the network to be a narrower term than \emph{rice}, helping both humans and machines in its interpretation.

\emph{Lexico-semantic diversity.} When an interlingual concept is lexicalized in a language (e.g.~\emph{rice} in French exists as \emph{riz}), it also appears as a \emph{language-specific concept} linked to it in the lexico-semantic layer of the lexicon. This allows characterising shared meanings by language-specific attributes, such as language-specific gloss, part of speech, etc.
Cases of lexical untranslatability (e.g.~no word exists for \emph{rice} in Swahili) are modelled by linking the interlingual concept to a \emph{lexical gap concept} instead, marked as GAP in the figure. The explicit distinction of untranslatability from words merely being absent from the lexicon---a frequent phenomenon due to the pervasive incompleteness of lexicons---can help MT systems identify difficult-to-translate phrases and handle them appropriately. 
A third kind of lexico-semantic diversity is modelled by language-specific \emph{local concepts} not merged into the interlingua. The Japanese lexicon in our example contains such a local concept for the Japanese word expressing \emph{raw brown rice}. Local concepts can form hierarchies, e.g.~\emph{raw brown rice} is a narrower term than \emph{raw rice}. Such hierarchies of local concepts are a key device in supporting linguistic diversity: through them, the UKC acknowledges the practical difficulty---if not impossibility---of integrating all culturally specific concepts from all societies into the single, coherent, global view of its interlingua. Even so, these local hierarchies remain connected to the interlingua through their root concepts, and can be exploited by applications destined to local communities. 

\emph{Lexical unity.} \emph{Rice}, \emph{riz}, and \emph{riso} do not only mean the same thing, they are also similar as word forms and are from a common etymological origin. As a unique feature among multilingual lexical databases, the UKC also models unity on the lexical level through \emph{cognate} relationships. Cognates are a key tool in linguistic typology and lexicostatistics for the study of the similarity of lexicons \cite{gudschinsky1956abc}. They are also used in cross-lingual NLP applications, such as for building bilingual word embeddings \cite{artetxe-etal-2016-learning}.

\emph{Lexical diversity.} Each language treats its words in its unique way. For example, Japanese and Italian both lexicalise \emph{rice in the husk} (also called \emph{paddy} in English), which therefore appear as word senses in the lexical layer of the UKC. Italian, however, expresses this concept through derivation, via the augmentative \emph{riso\,$\rightarrow$\,risone}. Morphological and semantic information that relate to the form of the word, such as derivation or antonymy relations between words, are modelled as lexical diversity in the UKC.


\begin{table}
    \setlength{\tabcolsep}{2pt}
    \centering
    \begin{tabular}{l|r|r|r}
    \textbf{Content type} & \textbf{Nodes~~~~~~~~~~~~~~~~~~~~} & \textbf{Relations} & \textbf{Langs covered} \\\hline
    Lex-Sem Unity & 111k concepts & 109k & 2,125 \\
    Lex-Sem Diversity & 1.5M concepts, 38k gaps & 13k & 2,125 \\
    Lexical Unity & -- & 3.5M & 764 \\
    Lexical Diversity & 2.8M word senses & 239k & 333 \\
    \end{tabular}
    \caption{Overview of UKC contents as of April 2023, in terms of lexico-\penalty-10000 semantic unity, lexico-semantic diversity, lexical unity, and lexical diversity.}
    \label{tab:ukc_contents}
\end{table}

As of early~2023, the UKC contains about 1.9~million words from over 2,100 languages.\footnote{The UKC does not contain named entities as it is not intended to be an encyclopedic resource nor to tackle issues typical of such resources.} 
The contents of the database in terms of  lexico-semantic and lexical unity and diversity are shown in Table~\ref{tab:ukc_contents}. 
Unity is expressed through 111k~interlingual concepts, 109k~lexico-semanic relations (hypernymy, meronymy, etc.), and 3.5M~cognate relations. Diversity is expressed via 38k~gaps from 744 languages, covering lexical domains such as family relationships and colours, 2.8M~word senses (i.e.~word--meaning pairs), and over 250k language-specific relations including derivation, antonymy, and metonymy.

Despite an underlying model designed to avoid structural bias in its interlingual mappings, and despite efforts to collect data on a wide range of languages, content-wise the UKC is not fully bias-free. First of all, it provides an uneven coverage of languages: for instance, only 7\% of the lexicons have more than 1000~words, while major European languages have lexicons of more than 50~thousand words. This situation reflects the general state of multilingual language resources (many of which the UKC also incorporates\footnote{See \cite{bella2022language}
for details about the resources from which the UKC was built.}). Secondly, the graph of interlingual concepts inside the UKC is not yet free from bias: having initially been bootstrapped from the concept hierarchy of the English Princeton WordNet, it continues to reflect, to some extent, an Anglo-Saxon point of view on the conceptualisation of the world, despite successful efforts in reorganising certain conceptual domains, such as kinship, according to a diversity-aware knowledge model \cite{khishigsuren2022using}.

\begin{figure}[t]
    \centering
    \includegraphics[width=.9\textwidth]{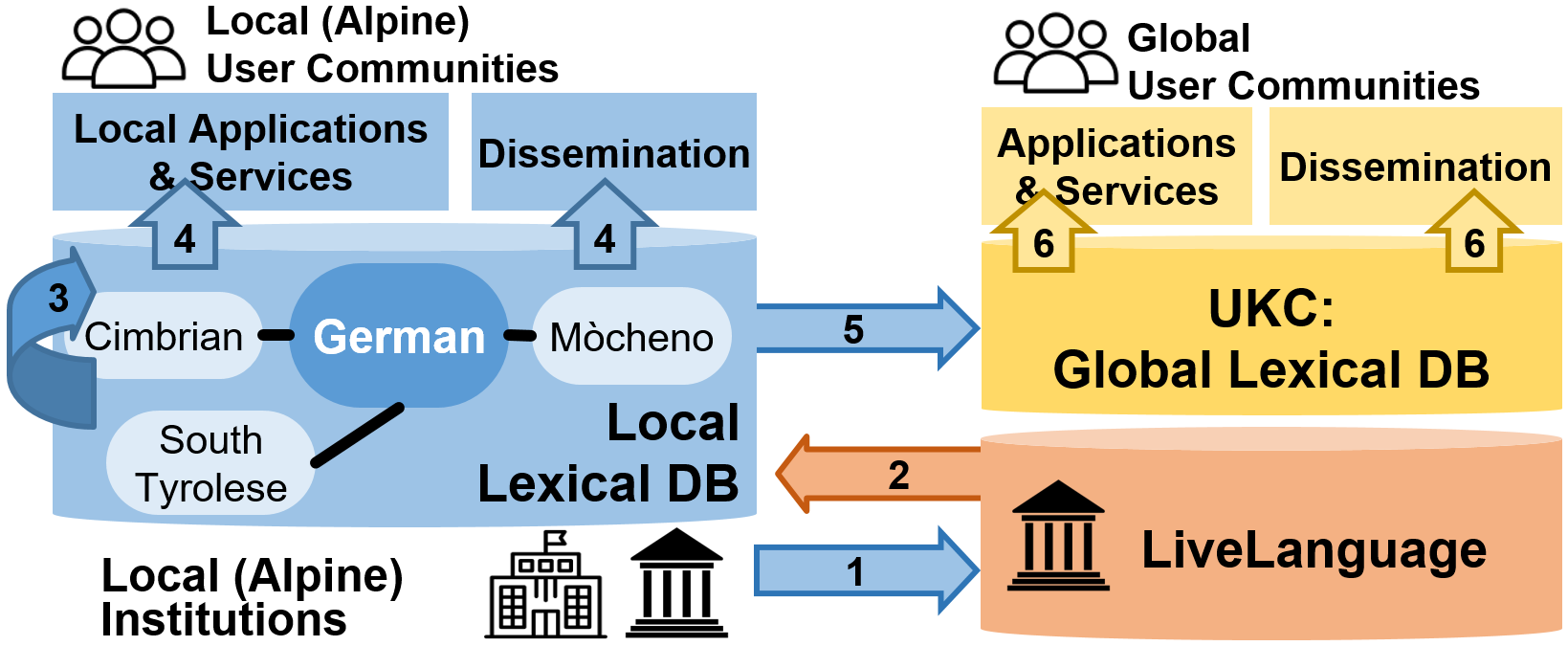}
    \caption{Collaborative language development methodology in the framework of the LiveLanguage initiative}
    \label{fig:multipolar}
\end{figure}

\section{A Diversity-Aware Development Methodology}
\label{sec:solution_methodology}

The aims of the \emph{LiveLanguage} initiative are to provide technical and methodological support for collaborative efforts on diversity-aware resource and tool development, 
and to disseminate the results of such efforts. According to the ethics policy of LiveLanguage, at the beginning of a development project, local communities get to decide on its goals and, at the end of the project, they keep the intellectual property of the results. These principles are an ethical minimum to avoid an exploitative relationship, but are also justified by efficiency: they ensure that the project is useful and relevant and motivate the local community in engaging with the project. A local institutional framework greatly simplifies the collaboration process: the local institution can act as the IP owner and has the necessary structure and network of people to organise the local effort.
Past and ongoing LiveLanguage projects have so far involved communities of Mongolia \cite{batsuren2019building}, Scotland \cite{bella2020major}, India \cite{chandran-nair-etal-2022-indoukc} and, more recently, the Middle-East, Indonesia, and South Africa. \footnote{See the complete list of projects on \url{http://ukc.datascientia.eu/projects}.} 
LiveLanguage adopts the following methodological steps for the co-creation of lexical resources, also depicted in Figure~\ref{fig:multipolar}:
\begin{enumerate}
    \item project specification based on local needs;
    \item local deployment of diversity-aware supporting tools;
    \item local resource development;
    \item local dissemination and exploitation of results;
    \item (optional) sharing of results with LiveLanguage;
    \item (optional) global dissemination and exploitation.
\end{enumerate}

\emph{Project specification} is determined by the long-term goals of the local community, and is led by the local institution. Goals can be as varied as language teaching, the development of AI-based language tech, basic language tools for smartphones, or the preservation of cultural heritage. Crucial design choices stem from these long-term goals, such as the lexical domain(s) or the trade and local language(s) or dialect(s) covered.
In line with the multipolar model of trade (hub) and local (satellite) languages, a hierarchical organisation of the project is also determined: a main local institution is charged with the project coordination, while its local partners (individuals or institutions) are in charge of efforts with respect to each satellite language. Let us take the example of a project on the languages and dialects within the Italian Alps, the goal of which is to expand existing lexicons for the purpose of teaching Alpine minority languages to local children. In this context, the trade languages are German and Italian, while the local languages and dialects can be the (Germanic) Mòcheno, South Tyrolean, and Cimbrian, or the (Romance) Ladin or Friulian. For each local language or dialect, the regional university coordinating the project takes care of involving   people (experts, speakers) or institutions from the respective communities.

\emph{Deployment of diversity-aware supporting tools.} Upon request, LiveLanguage can provide software and hardware infrastructure (servers, tools, data) to the local institution, as well as training and consultancy for the development process. In Figure~\ref{fig:multipolar}, for an Alpine project, a \emph{Local Lexical DB} is generated with the German lexicon as a trade language, and the existing (preliminary and incomplete) lexicons of three local languages, all automatically downloaded from the UKC via the LiveLanguage data catalogue. At the current stage, LiveLanguage provides the following software support:
\begin{itemize}
    \item download of individual or interconnected multilingual lexicons in a standard format from the LiveLanguage data catalogue;\footnote{\url{http://livelanguage.eu}.}
    \item a simple-to-use, open source lexical DB management system, automatically preloaded with existing UKC lexical data on the hub and satellite languages, mapped together as illustrated in Figure~\ref{fig:mldb_mapping};
    \item browsing and visualisation tools for the multilingual lexicons, such as local versions of the UKC website\footnote{See, for example, \url{http://indo.ukc.datascientia.eu}.};
    \item tools for the editing of lexicons.
\end{itemize}

\emph{Local development, dissemination, and exploitation.} The local institution(s) manage the process of resource development. They involve local collaborators according to project needs and, if necessary, may ask for consultancy (typically free of charge) from LiveLanguage. As they get to keep intellectual property rights over the resources produced, they have freedom to define the IP policies that govern the use of results, as well as to disseminate or exploit them through local applications or services. LiveLanguage provides tools both for development and dissemination of results.

\emph{Global integration and dissemination.} Local institutions are encouraged to share project results with LiveLanguage in their own interest: LiveLanguage offers the added value of mapping the local lexicons, by means of the UKC as a global lexical database, to all other languages. In other words, local efforts provide the element of \emph{diversity} into the lexicon, and integration with the UKC provides interlingual \emph{unity}. The appearance of local lexicons in the UKC and the LiveLanguage data catalogue also provides an additional dissemination opportunity for the effort.     

\section{Conclusions and Future Work}

While recognising that technology with a completely bias-free representation of the world's languages is an unattainable ideal, our case study demonstrates how a language resource development project can take concrete steps in that direction. The UKC and the LiveLanguage initiative are long-term projects that address both linguistic diversity and linguistic bias on the technological, methodological, ethical, practical, and social levels.
The UKC was presented to the global public in 2021 and has since been expanded with large amounts of lexical data on diversity. The collection of such data is a never-ending challenge, and we are initiating more projects, as well as offering new tools and services in the near future.

Lastly, the collaborative process described in the previous section implicitly assumes the existence of a central organisation taking care of the long-term maintenance and sustainability of key LiveLanguage components: the UKC database and website, the LiveLanguage data catalogue, as well as the tools and services. While currently the University of Trento is playing this role, our short-term plans involve the creation of the DataScientia Foundation\footnote{\url{http://datascientia.eu}.} 
as a not-for-profit coordinating body where diverse stakeholders share decisional and operational power, incuding an international advisory board featuring experts from various language communities.

\bibliographystyle{main}
\bibliography{main}
\end{document}